\documentclass[journal,onecolumn]{IEEEtran}

\usepackage{epsfig}
\usepackage{graphicx}
\usepackage{amsmath}
\usepackage{amssymb}
\usepackage{commath}
\usepackage{bbding}
\usepackage{cite}
\usepackage{cleveref}
\usepackage{booktabs,multirow}
\usepackage{times}

\usepackage{mathrsfs}

\usepackage{mathtools}

\newtheorem{definition}{Definition}

\DeclareMathOperator*{\med}{Median}

\begin{document}

\title{An Invariant Model of the Significance of Different Body Parts in Recognizing Different Actions}
%
%

\author{Yuping~Shen\thanks{Yuping Shen was with the Department of Computer Science, University of Central Florida, Orlando, FL, 32816 USA at the time this project was conducted. (e-mail: 
ypshen@cs.ucf.edu).} and Hassan~Foroosh\thanks{Hassan Foroosh is with the Department of Computer Science, University of Central Florida, Orlando, FL, 32816 USA (e-mail: foroosh@cs.ucf.edu).}
}

\maketitle

\begin{abstract}
In this paper, we show that different body parts do not play equally important roles in recognizing a human action in video data. We investigate to what extent a body part plays a role in recognition of different actions and hence propose a generic method of assigning weights to different body points. The approach is inspired by the strong evidence in the applied perception community that humans perform recognition in a foveated manner, that is they recognize events or objects by only focusing on visually significant aspects. An important contribution of our method is that the computation of the weights assigned to body parts is invariant to viewing directions and camera parameters in the input data. We have performed extensive experiments to validate the proposed approach and demonstrate its significance. In particular, results show that considerable improvement in performance is gained by taking into account the relative importance of different body parts as defined by our approach.
\end{abstract}


\section{Introduction}

Human action recognition from video data data has a wide range of applications in areas such as surveillance and image retrieval \cite{Junejo_Foroosh_2008,Sun_etal_2012,junejo2007trajectory,sun2011motion,Ashraf_etal2012,sun2014feature,Junejo_Foroosh2007-1,Junejo_Foroosh2007-2,Junejo_Foroosh2007-3,ashraf2012motion,ashraf2015motion,sun2014should}, image annotation \cite{Tariq_etal_2017,Tariq_etal_2017_2,tariq2013exploiting,tariq2015feature,tariq2014scene}, video post-production and editing \cite{balci2006real,moore2008learning,alnasser2006image,Alnasser_Foroosh_rend2006,fu2004expression,balci2006image},
and self-localization \cite{Junejo_etal_2010,Junejo_Foroosh_2010,Junejo_Foroosh_solar2008,Junejo_Foroosh_GPS2008,junejo2008gps}, to name a few.

The literature on human action recognition from video data includes both monocular and multiple view methods \cite{Shen_Foroosh_2009,Ashraf_etal_2014,Sun_etal_2015,shen2008view,sun2011action,ashraf2014view,shen2008action,shen2008view-2,ashraf2010view,boyraz122014action,Shen_Foroosh_FR2008,Shen_Foroosh_pose2008,ashraf2012human}. Often, multiple view methods are designed to tackle viewpoint invariant recognition \cite{Shen_Foroosh_2009,Ashraf_etal_2014,shen2008view,ashraf2014view,shen2008view-2,ashraf2010view,Shen_Foroosh_FR2008,ashraf2012human}, although such methods may require calibration across views \cite{Junejo_etal_2011,junejo2006dissecting,junejo2007robust,junejo2006robust,Junejo_Foroosh_calib2008,Junejo_Foroosh_PTZ2008,Junejo_Foroosh_SolCalib2008,Ashraf_Foroosh_2008,Junejo_Foroosh_Givens2008,Balci_Foroosh_metro2005},  image registration \cite{Foroosh_etal_2002,Foroosh_2005,Balci_Foroosh_2006,Balci_Foroosh_2006_2,Alnasser_Foroosh_2008,Atalay_Foroosh_2017,Atalay_Foroosh_2017-2,shekarforoush1996subpixel,foroosh2004sub,shekarforoush1995subpixel,balci2005inferring,balci2005estimating,Balci_Foroosh_phase2005,Foroosh_Balci_2004,foroosh2001closed,shekarforoush2000multifractal,balci2006subpixel,balci2006alignment,foroosh2004adaptive,foroosh2003adaptive}, or tracking across views \cite{Shu_etal_2016,Milikan_etal_2017,Millikan_etal2015,shekarforoush2000multi,millikan2015initialized}. There are also methods that rely on human-object interaction \cite{prest2012weakly,yao2011human,yao2012recognizing}, which often require identifying image contents other than humans \cite{liu2015sparse,wang2016factorized,Cakmakci_etal_2008,Cakmakci_etal_2008_2,Lotfian_Foroosh_2017,Morley_Foroosh2017,Ali-Foroosh2016,Ali-Foroosh2015,Einsele_Foroosh_2015,ali2016character,Cakmakci_etal2008,damkjer2014mesh}.
Other preprocessing steps that may be needed include image restoration \cite{Foroosh_2000,Foroosh_Chellappa_1999,Foroosh_etal_1996,berthod1994reconstruction,shekarforoush19953d,lorette1997super,shekarforoush1998multi,shekarforoush1996super,shekarforoush1995sub,shekarforoush1999conditioning,shekarforoush1998adaptive,berthod1994refining,shekarforoush1998denoising,bhutta2006blind,jain2008super,shekarforoush2000noise,shekarforoush1999super,shekarforoush1998blind},
or scene modeling \cite{Junejo_etal_2013,bhutta2011selective,junejo1dynamic,ashraf2007near}.

In this paper, we look at a very specific problem of determining which body parts play a role in recognizing an action and to what extent. This is a crucial information that may serve as a prior information to any method in the literature, whether video-based or single-image method.

\section{Related Work}\label{sec:bgActionRec}
The literature in human action recognition has been extremely active in the past two decades and significant progress has been made in this area \cite{gavrila1999vah, moeslund2001scv, moeslund2006sav, wang2003rdh}. Human action recognition methods start by assuming a model of the human body, e.g. silhouette, body points, stick model, etc., and build algorithms that use the adopted model to recognize body pose and its motion over time. While there has been many different ways of classifying existing action recognition methods in the literature \cite{gavrila1999vah, moeslund2001scv, moeslund2006sav, wang2003rdh}, we present herein a different perspective. We consider action recognition at three different levels: (i) the subject level, which requires studying the kinematics of the articulated human body, (ii) the image level, which requires studying the process of imaging the 3D  human body both in terms of the geometry and lighting effects, and (iii) the observer level, which requires studying how an observer would interpret visual data and recognize an action. This way of categorizing action recognition is rather uncommon but offers a different point of view of how one can approach this problem.

Herein we are interested in investigating observer level issues from a new point of view. It has been long argued in the applied perception community \cite{Schutz2009} that human observers use a foveated strategy, which may be interpreted as an approach using ``importance sampling''. What this implies is that humans sample only the most significant aspects of an event or action for recognition, and do not give equal importance to every observed data point. Most of existing action recognition methods in the literature have primarily focused on subject level \cite{Blank2005,bb53505,bobick2001rhm,wang2003sab,yilmaz2005asn,parameswaran2003vih,yilmaz2006map} and image level \cite{efros2003rad,laptev2005pmd,schuldt2004rha,zhu2006arb,wang2006awg} issues. A number of methods have focused on the observer level \cite{ahmad2006hbh,schuldt2004rha}, but mostly from a machine learning point of view. An interesting work by Sheikh et al. \cite{sheikh2005esh} recently has perhaps a close connection to what we aim to address in this paper. However, they investigate a holistic approach in an abstract framework of action space. Our goal is to propose an observer level approach with more direct connection to image level features, which we believe would be more ``intuitive''. In other words, we are interested in exploring an idea similar to importance sampling where image level features are directly assigned different importance weights that are derived directly from the extent to which the feature affects the recognition performance for a given action.

\begin{figure}[htb]
\centering
\includegraphics[width=70mm]{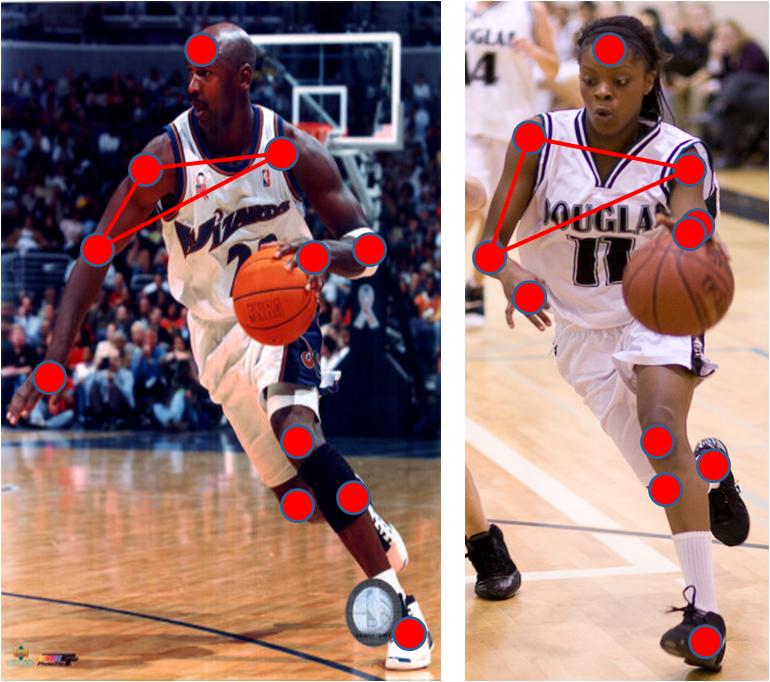}
\caption{An example of human poses, with the points used for body representation: head, shoulders, elbows, hands, knees, and feet. A body point triplet is shown.}
\label{fig:body}
\end{figure}

One particular issue that has attracted attention among many of these groups is invariance \cite{campbell1996ifg,gritai2004uai,junejo2008cross,parameswaran2003vih,rao2002vir,Shen2008,Shen2009,rao2002vir}. Invariance is an important issue in any recognition problem from image or video data, since appearance is substantially affected by the imaging level, i.e. the viewing angle, camera parameters, lighting, etc. In action recognition the issue of invariant recognition is compounded with additional complexity introduced by the high degrees of freedom of human body \cite{Zatsiorsky02}.

Our focus in this paper is twofold: to investigate action recognition at the observer level which takes into account how humans recognize actions by paying attention to only the most significant visual cues, and to introduce this approach within an invariant framework

\section{Our Body Model}

We use a point-based model similar to \cite{Shen2008,Shen2009}, which consists of a set of 11 body points (see Figure \ref{fig:body}). We also adopt the idea of decomposing the non-rigid motion of human body into rigid motions of planes given by triplets of body points \cite{Shen2008,Shen2009}. This implies that the non-rigid motion of human body is described by a set of homographies induced by planes associated with the body point triplets . As a result the complex problem of estimating the human body motion is reduced to a set of linear problems associated with the motions of planes.

\begin{figure}[htb]
\centering
\begin{tabular}{c}
\includegraphics[width=7in]{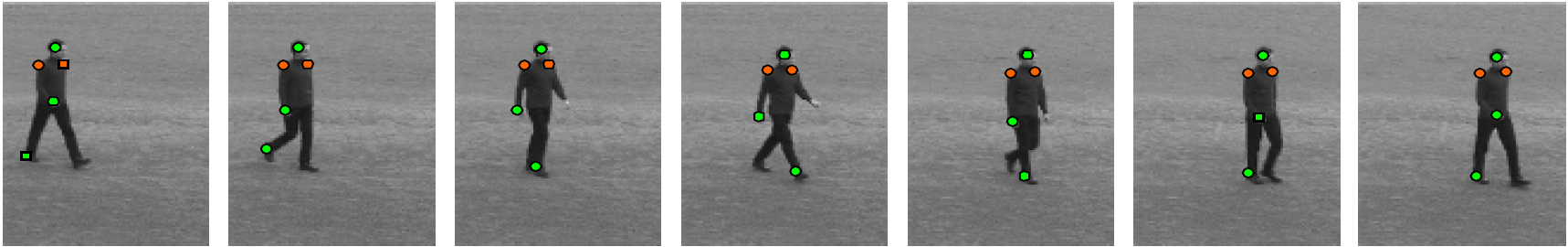}\\
\includegraphics[width=7in]{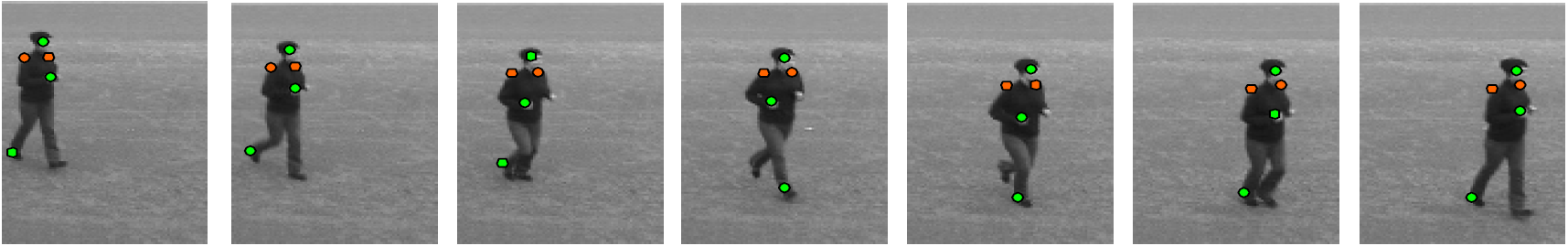}\\
(a) Examples of walking (upper sequence) and jogging (lower sequence)\\
\includegraphics[width=7in,height=1.5in]{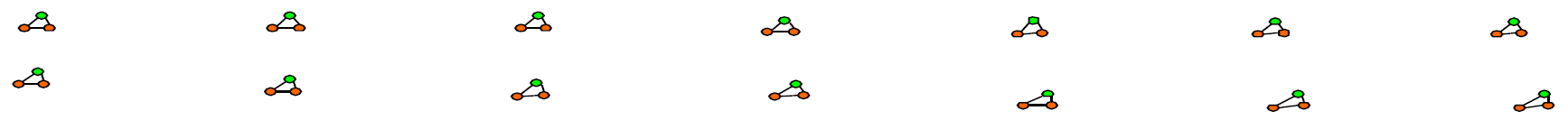}\\
(b) Comparison of head-shoulders triplet in walking and jogging\\
\includegraphics[width=7in]{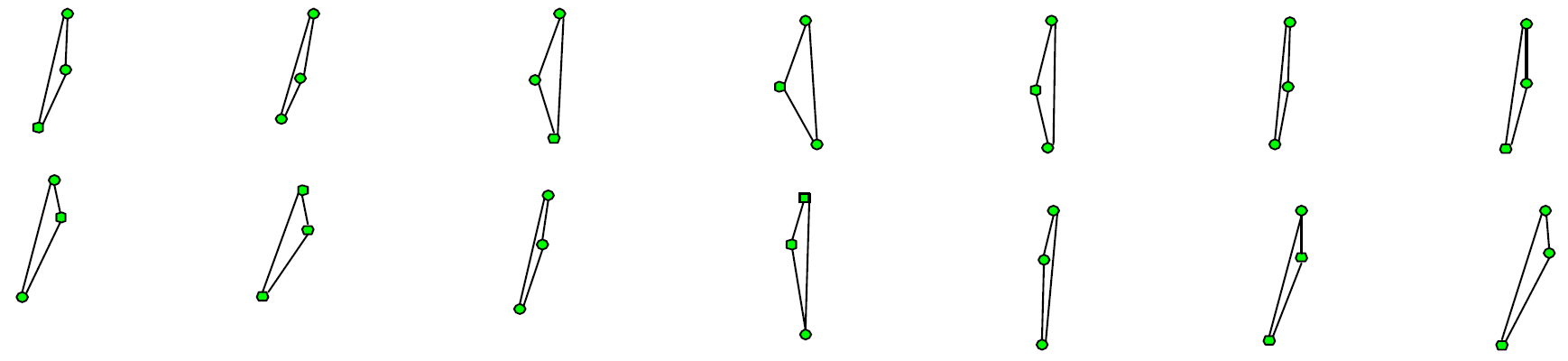}\\
(c) Comparison of head-hand-knee triplet in walking and jogging
\end{tabular}
\caption{Roles of different triplets in action recognition}
\label{fig:walkRunDiff}
\end{figure}

With the triplet representation of human pose and action, we may consider the relative importance of body point triplets for different actions. For instance, the triplets composed of shoulders and hips have similar motion in walking and jogging, and thus make trivial contribution to distinguish them, while other triplets that consist of shoulder, knee and foot joints carry more essential information of the differences between walking and jogging (See Figure \ref{fig:walkRunDiff}). Understanding the roles of body point triplets in human motion/action could help us retrieve more accurate information on human motion, and thus improve the performance of our recognition methods.

Our problem can be described as follows. We have a database of reference actions. Given a target sequence we are interested in finding out which reference sequence in the database best matches our target sequence. We are interested in performing this task in a manner that is invariant to viewing directions and the camera parameters. Our first step is to align the target sequences to be recognized to the reference sequences in our database, as described in the next section.

\subsection{Sequence Alignment}
The body motion perceived in two different frames is often referred to as body pose transition. Our sequence alignment method is based on finding corresponding pose transitions in two actions sequences viewed by two different cameras. Our starting point is the view-invariant similarity measure that was proposed in \cite{Shen2008,Shen2009}: Any triplet of body points in one camera iamge and the corresponding triplet of points in a second camera define a homography ${\bf H}_1$ between the two cameras. After a pose transition the triplets move to new positions defining a new homography ${\bf H}_2$. If the triplet motions are similar, then the two homographies become consistent with the fundamental matrix, and as a result the homography defined by ${\bf H} = {\bf H}_1{\bf H}_2^{-1}$ will become a homology, two of whose eigenvalues are equal. The equality of the two eigenvalues of the homography ${\bf H}$, provides thus a measure of similarity for the motion of the two point triplets, which is invariant to viewing angles and the camera parameters \cite{Shen2008}. Given $11$ body points as shown in Figure 1, there are $165$ such triplets, all of which are used to provide a measure of similarity for the two actions being compared. We use the following measure of similarity over all possible triplets for evaluating the similarity of two pose transitions $\mathcal{T}_1$ and $\mathcal{T}_2$:

\begin{eqnarray}
\mathscr{S}(\mathcal{T}_1,\mathcal{T}_2) &=& \sum_{i}\frac{\left|a_i-b_i\right|}{\left|a_i+b_i\right|} \label{eq:costf2}
\end{eqnarray}
where $a_i$ and $b_i$ are the two closest eigenvaluse of the homography defined by the $i$-th triplet, and $i=1,...,165$ for 11 body points. \\

Given a target sequence of $m$ pose transitions $\mathcal{A}_j$, $j=1,...,m$ and a reference sequence of $n$ pose transitions $\mathcal{B}_{j'}$, $j'=1,...,n$, in order to find the optimal alignment $\psi: \mathcal{A}\rightarrow \mathcal{B}$ between the two actions $\mathcal{A}$ and $\mathcal{B}$, we build a matching error matrix using (ref{eq:costf2}), and use dynamic programming to find the optimal mapping.
Once two sequences are aligned, the main issue is to determine how similar the two actions are. In the next section, we describe our weighting-based action recognition method.

\section{Weighting-based Human Action Recognition}\label{sec:weightedActRec}
To study the roles of body-point triplets in action recognition, we select two different sequences of walking action $WA = \{I_{1\ldots l}\}$ and $WB = \{J_{1\ldots m}\}$, and a sequence of running action $R = \{K_{1\ldots n}\}$. We then align sequence $WB$ and $R$ to $WA$, using the alignment method described in Section 2, and obtain the corresponding alignment/mapping $\psi: WA \rightarrow WB$ and $\psi' : WA \rightarrow R$. As discussed in Section 2, the similarity of two poses is computed based on error scores of all body-point triplets motion. For each matched poses $\left<I_i, J_{\psi(i)}\right>$, we stack the error scores of all triplets as a vector $\mathbf{V}_{e}(i)$:
\begin{equation}
\label{eqn:EVi}
\mathbf{V}_{e}(i) =
 \left[\begin{array}{c}
E(\Delta_1)\\
E(\Delta_2)\\
\colon\\
E(\Delta_T)
\end{array}\right],
\end{equation}
where $E(\Delta_i)=\frac{\left|a_i-b_i\right|}{\left|a_i+b_i\right|}$.\\

\begin{figure}[htb]
\centering
\includegraphics[width=60mm,height=3in]{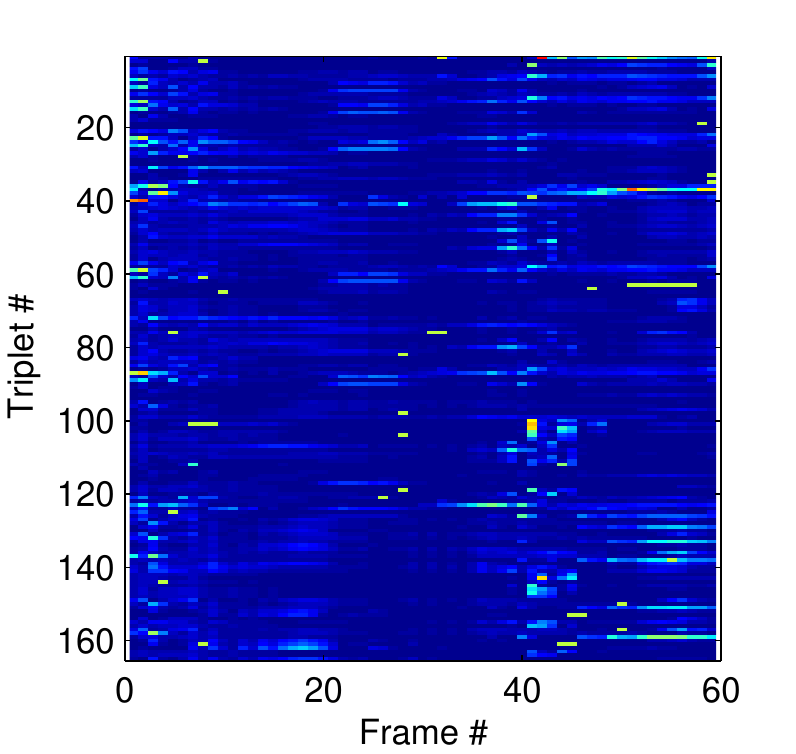} \includegraphics[width=60mm,height=3in]{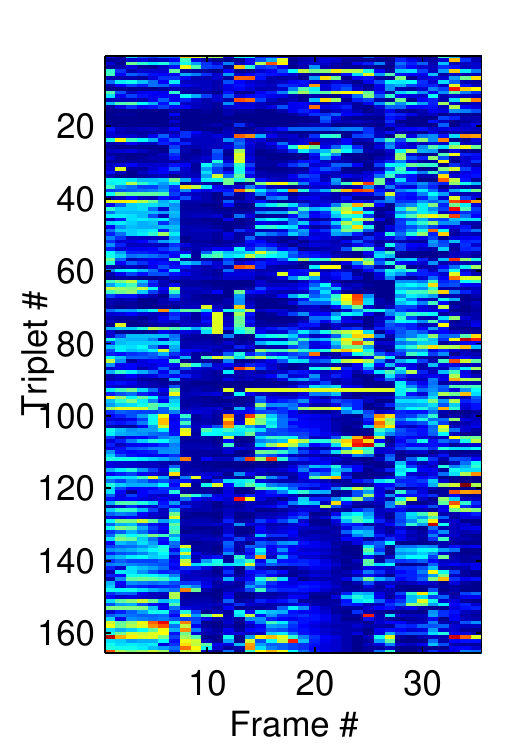}
\caption{Visual illustration of $\mathbf{M}_e$ (left) and $\mathbf{M}'_e$ (right)}
\label{fig:compareDiffActions}
\end{figure}

We then build an error score matrix $\mathbf{M}_e$ for alignment $\psi_{WA \rightarrow WB}$:
\begin{equation}
\mathbf{M}_e = \left[\begin{array}{cccc}
\mathbf{V}_{e}(1) & \mathbf{V}_{e}(2) & \ldots & \mathbf{V}_{e}(l)
\end{array}\right].
\end{equation}
Each row $i$ of $\mathbf{M}_e$ indicates the dissimilarity scores of triplet $i$ across the sequence, and the expected value of each column $j$ of $\mathbf{M}_e$ is the dissimilarity score of pose $I_j$ and $J_{\psi_{WA \rightarrow WB}(j)}$. Similarly we build an error score matrix $\mathbf{M}'_e$ for alignment $\psi_{WA \rightarrow R}$. $\mathbf{M}_e$ and $\mathbf{M}'_e$ are illustrated visually in Figure \ref{fig:compareDiffActions}.

\begin{figure*}[htb]
\centering
\begin{tabular}{ccc}
\hspace*{-0.3cm}\includegraphics[width=60mm]{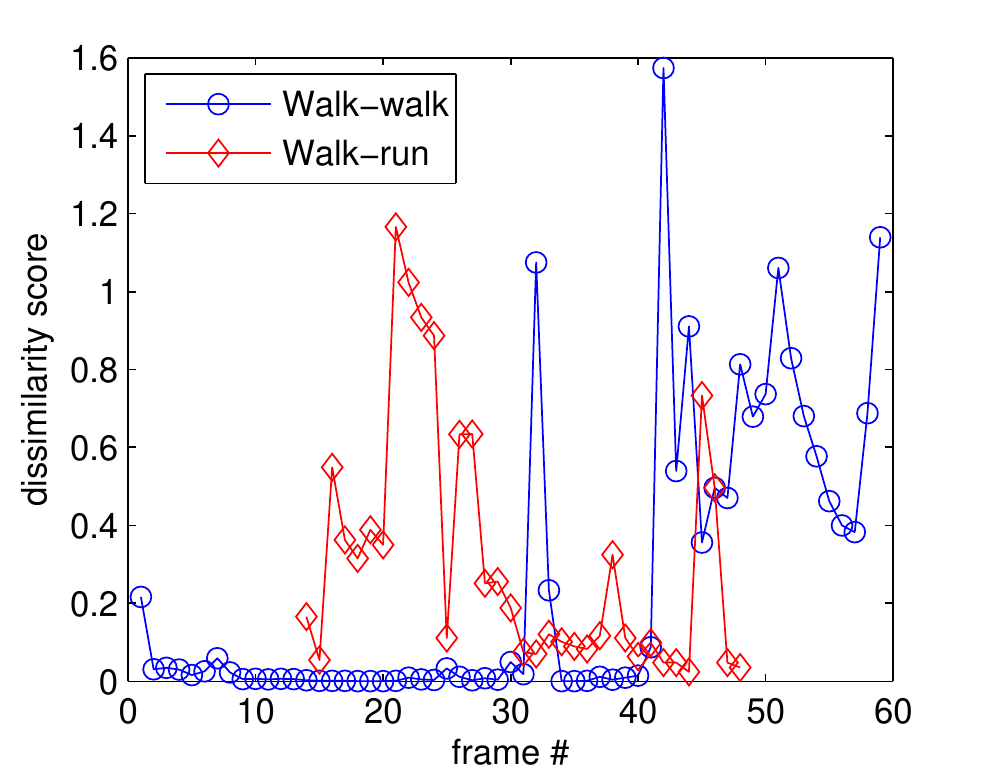} &\hspace*{-0.3cm}\includegraphics[width=60mm]{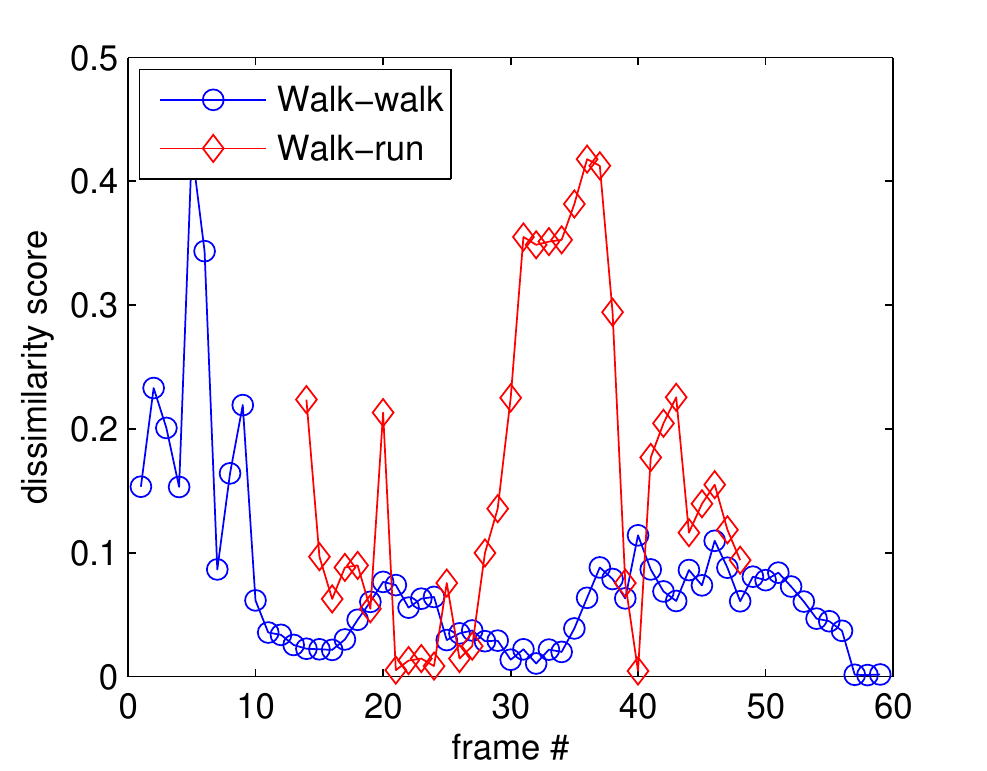} & \hspace*{-0.3cm}\includegraphics[width=60mm]{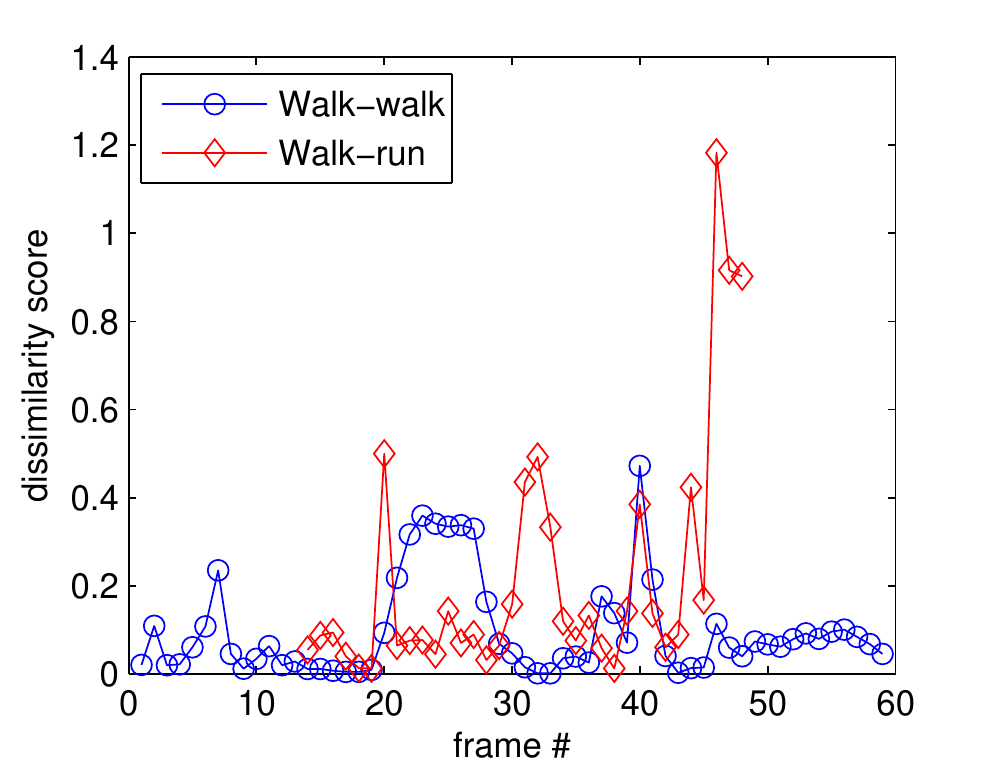} \\
(a) Triplet 1 & (b) Triplet 21 &(c) Triplet 90\\
\hspace*{-0.3cm}\includegraphics[width=60mm]{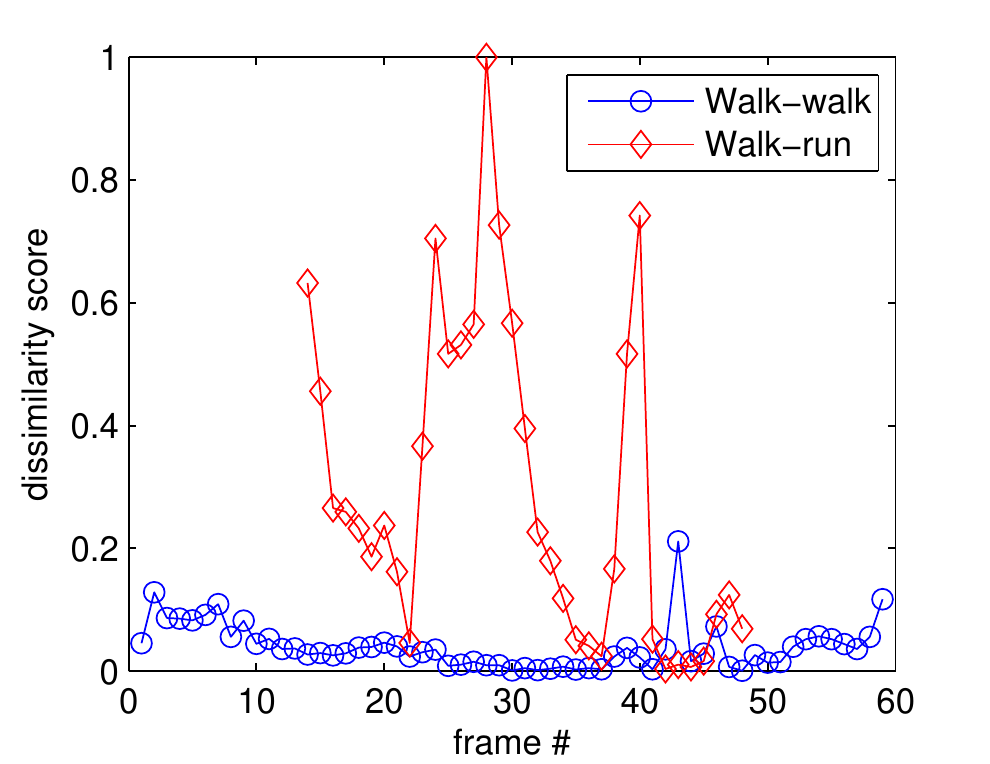} &\hspace*{-0.3cm}\includegraphics[width=60mm]{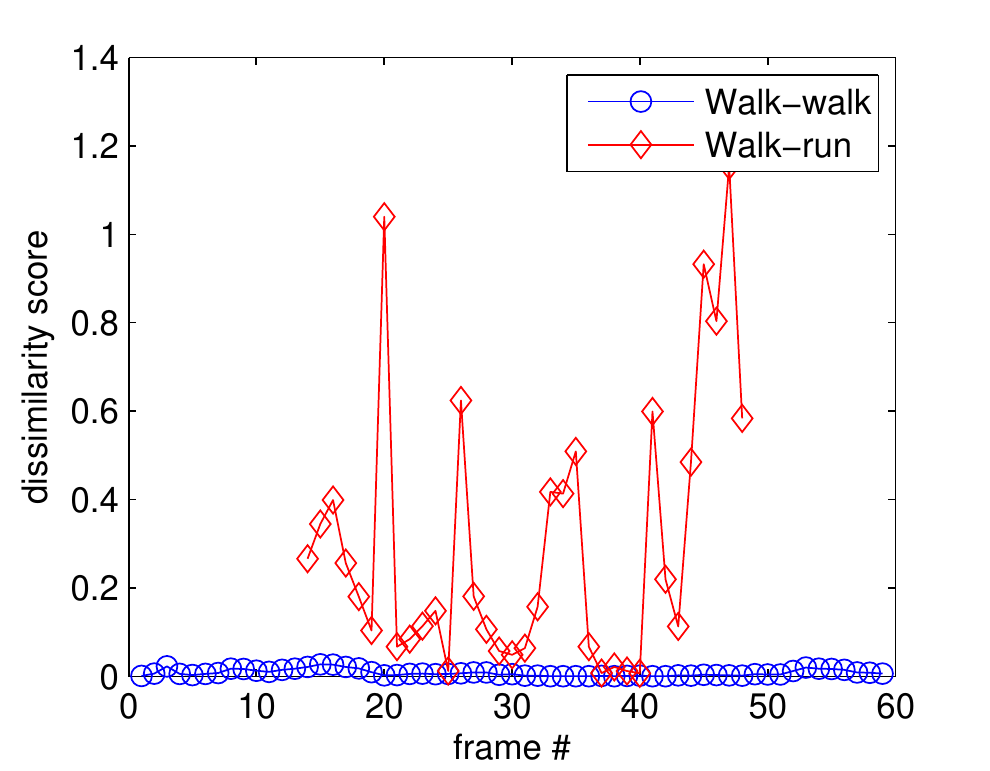} & \hspace*{-0.3cm}\includegraphics[width=60mm]{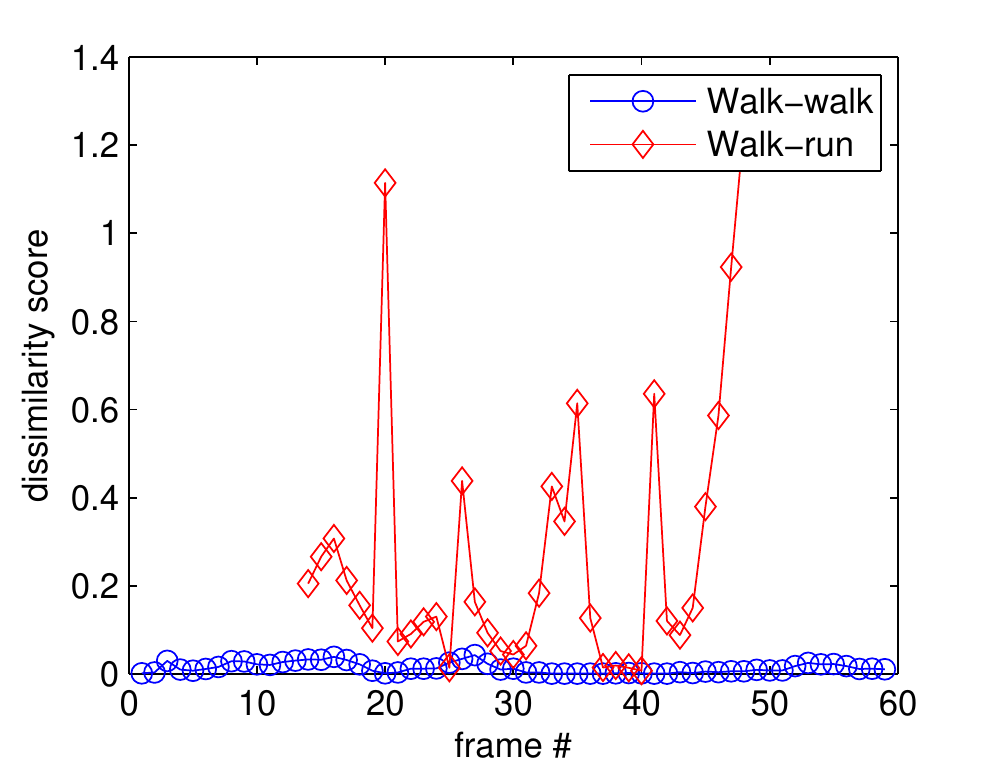} \\
(d) Triplet 55 &(e) Triplet 94 & (f) Triplet 116
\end{tabular}
\caption{Roles of different triplets in action recognition. (a) - (f) are the plots of dissimilarity scores of some triplets across frames in the walk-walk and walk-run alignments.}
\label{fig:compareDiffActions2}
\end{figure*}

To study the role of a triplet $i$ in distinguishing walking and running, we compare the $i$-th row of $\mathbf{M}_e$ and $\mathbf{M}'_e$, as plotted in Figure \ref{fig:compareDiffActions2} (a) - (f). We found that, some triplets such as triplets 1 , 21 and 90 have similar error scores in both cases, which means the motion of these triplets are similar in walking and running. On the other hand, triplets 55, 94 and 116 have high error scores in $\mathbf{M}'_e$ and low error scores in $\mathbf{M}_e$, that is, the motion of these triplets in a running sequence is different from their motion in a walking sequence. Triplets 55, 94 and 116 reflect the variation in actions of walking and running, thus are more informative than triplets 1 , 21 and 90 for the task of distinguishing walking and running actions.

\begin{figure*}[htb]
\centering
\begin{tabular}{ccc}
\hspace*{-0.3cm}\includegraphics[width=60mm]{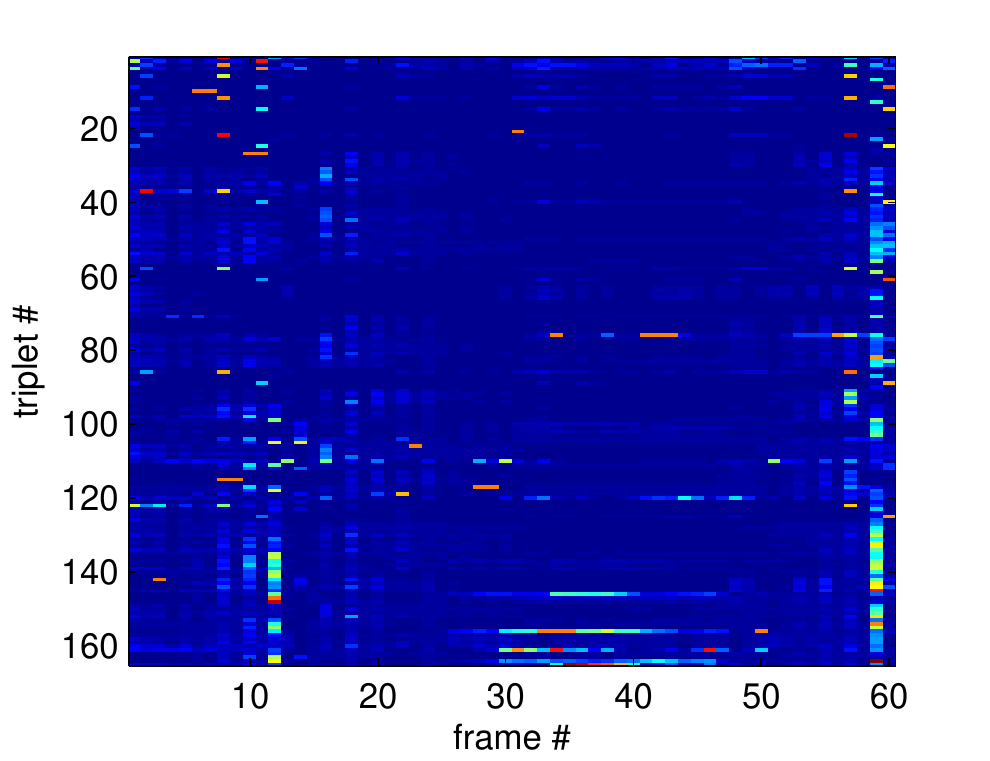} &\hspace*{-0.3cm}\includegraphics[width=60mm]{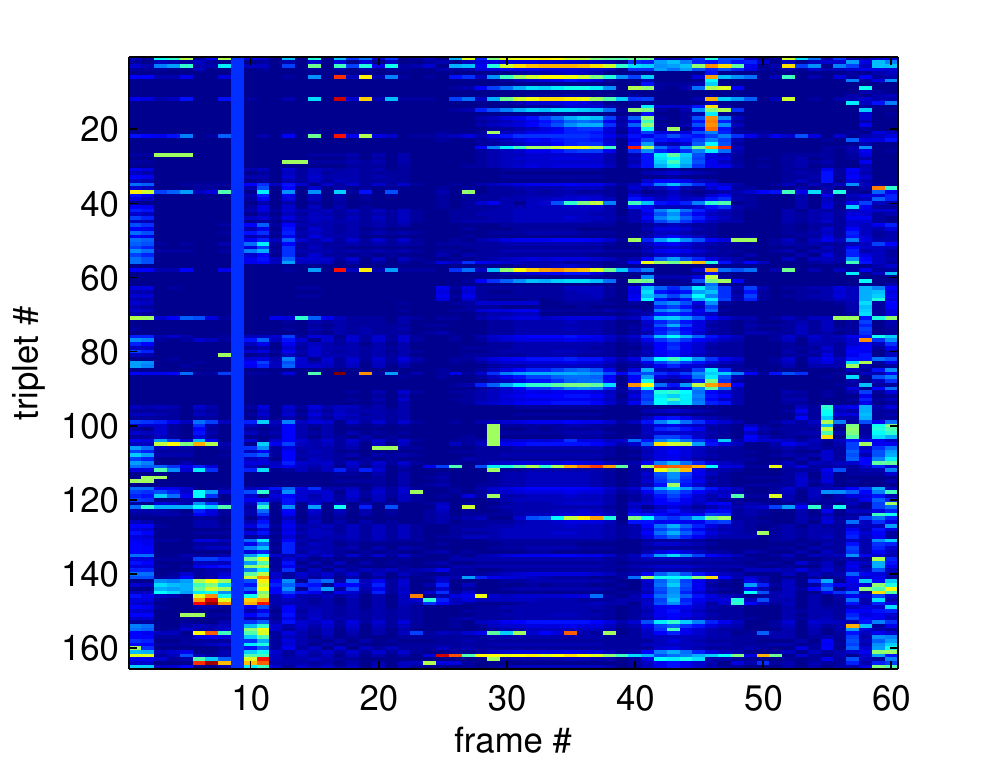} & \hspace*{-0.3cm}\includegraphics[width=60mm]{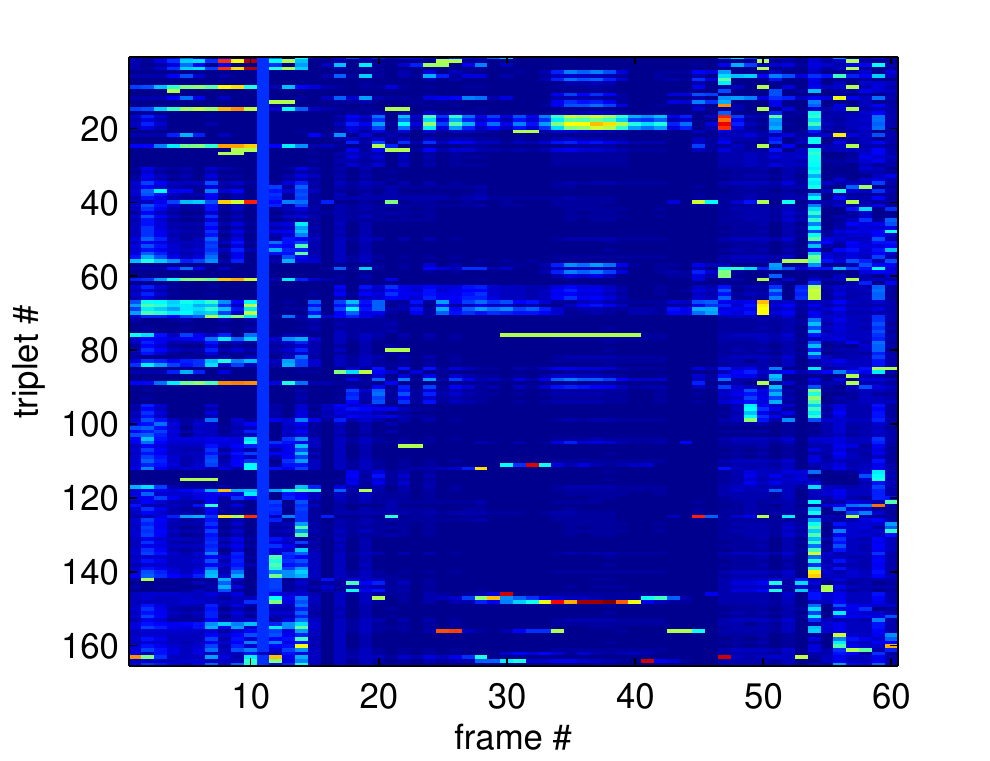} \\
(a) $\mathbf{M}^1_e$ & (b) $\mathbf{M}^2_e$ & (c) $\mathbf{M}^3_e$\\
\hspace*{-0.3cm}\includegraphics[width=60mm]{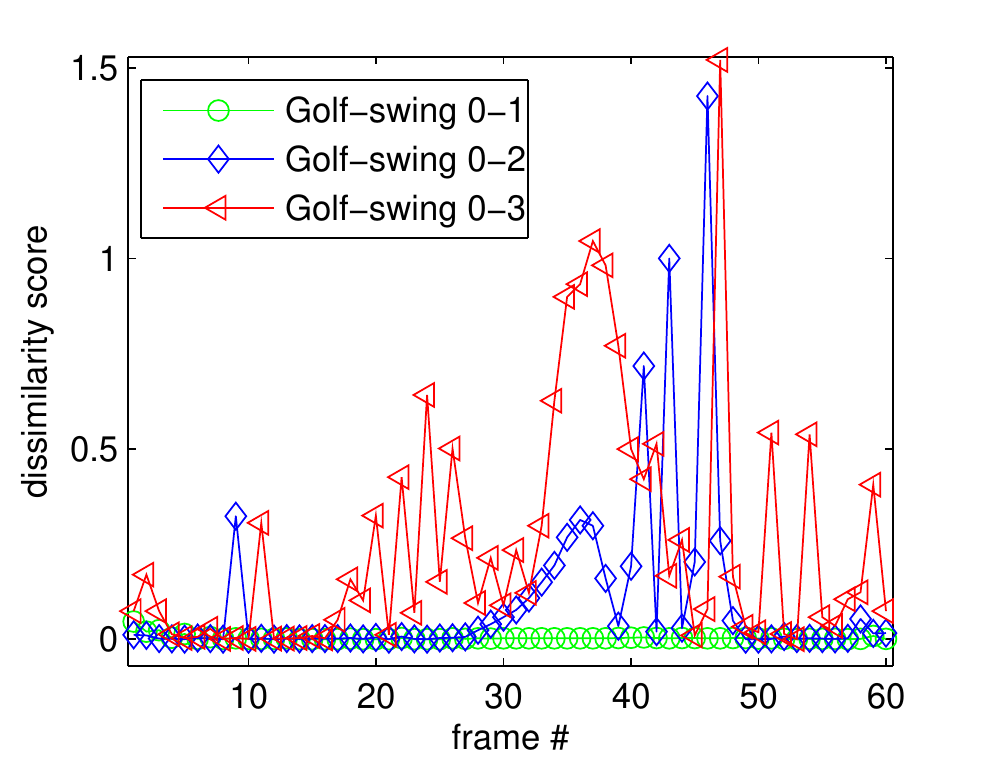} &\hspace*{-0.3cm}\includegraphics[width=60mm]{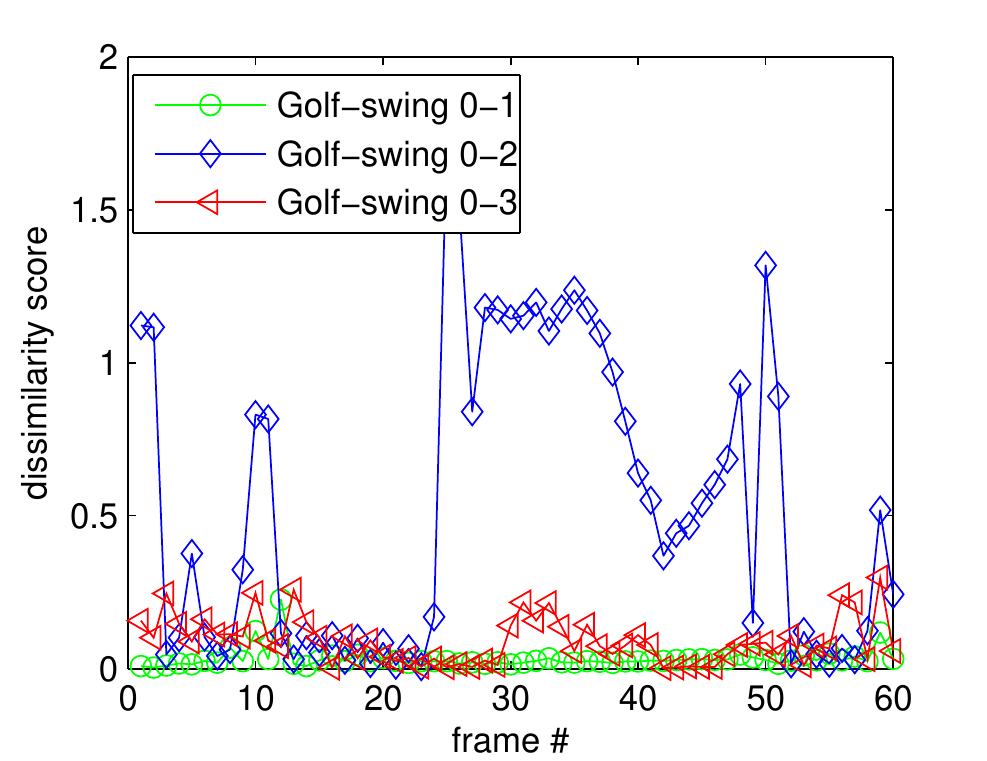} & \hspace*{-0.3cm}\includegraphics[width=60mm]{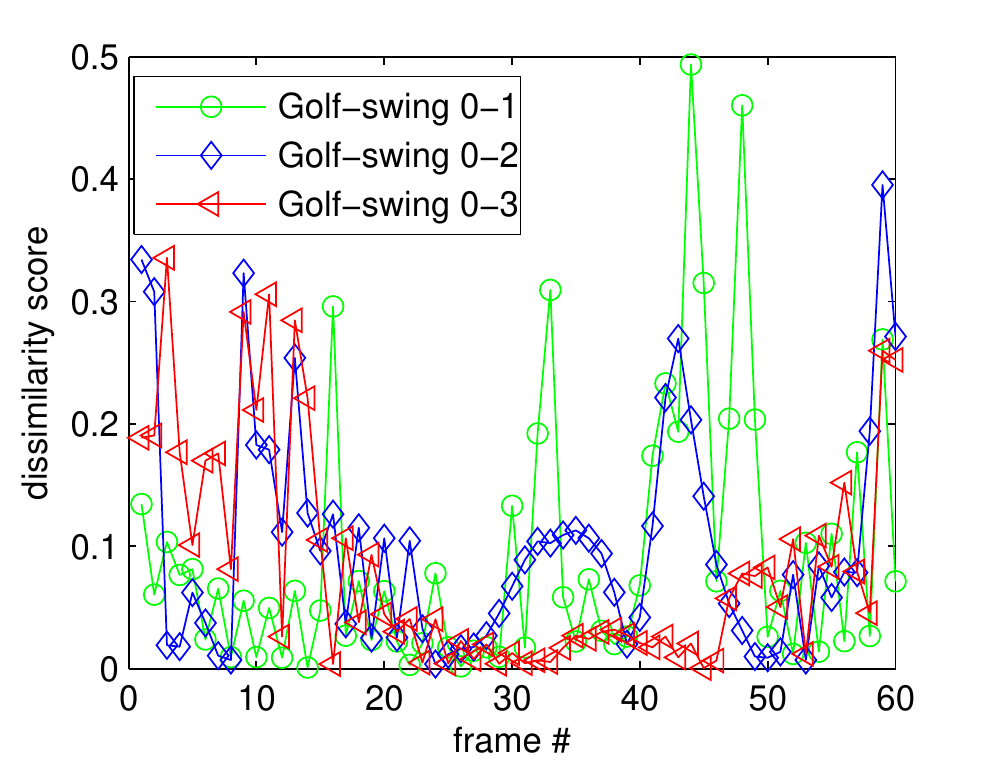} \\
(d) Triplet 20 &(e) Triplet 162 & (f) Triplet 20
\end{tabular}
\caption{Roles of different triplets in action recognition}
\label{fig:compareSameActions}
\end{figure*}

In the following experiments, we compare sequences of different individuals performing the same action, and study the roles of triplets in categorizing them in the same group of action: Select four sequences $G0$, $G1$, $G2$, and $G3$ of golf-swing action, and align $G1$, $G2$, and $G3$ to $G0$ using the alignment method described in Section 2, and then build error score matrix $\mathbf{M}^1_e$, $\mathbf{M}^2_e$, $\mathbf{M}^3_e$ correspondingly as in above experiments. From the illustrations of $\mathbf{M}^1_e$, $\mathbf{M}^2_e$, $\mathbf{M}^3_e$ in Figure \ref{fig:compareSameActions} (a), (b) and (c). The dissimilarity scores of some triplets, such as triplet 120 (see  Figure \ref{fig:compareSameActions} (f)) , is very consistent across individuals. Some other triplets such as triplets 20 (Figure \ref{fig:compareSameActions} (d)) and 162 (Figure \ref{fig:compareSameActions} (e)) have various error score patterns across individuals, that is, these triplets represent the variations of individuals performing the same action.

\begin{definition}\label{def:essentialtriplet}
If a triplet reflects the essential differences between an action $A$ and other actions, we call it a {\textbf significant triplet} of action $A$. All triplets other than significant triplets are referred to as {\textbf trivial triplets} of action $A$.
\end{definition}

A typical significant triplet should (1) convey the variations between actions and/or (2) tolerate the individual variations of the same action.
For example, triplets 55, 94 and 116 are significant triplets for walking action, and triplet 20 is a significant triplet for the golf-swing action.

Intuitively, in the task of action recognition, we should place more focus on the significant triplets while reducing the negative impact of trivial triplets, that is, assigning appropriate influence factor to the body-point triplets. In our approach to action recognition, this can be achieved by assigning appropriate weights to the similarity errors of body point triplets in equation (1). That is, equation (1) could be rewritten as:
\begin{equation}\label{eqn:weightedSimilarity}
\mathcal{E}(I_1\rightarrow I_2, J_i\rightarrow J_j) = \sum_{\mbox{all } \Delta_i}(\lambda_i \cdot E(\Delta_i)),
\end{equation}
where
$\lambda_1 + \lambda_2 + \ldots +\lambda_T = 1, T = \binom{n}{3}$, $n$ is the number of body points in the human body model.

The next question is, how to determine the optimal set of weights $\lambda_i$ for different actions. Manual assignment of weights could be biased and difficult for a large database of actions, and is inefficient when new actions are added in. Automatic assignment of weight values is desired for a robust and efficient action recognition system. To achieve this goal, we propose to use a fixed size dataset of training sequences to learn weight values. Suppose we are given a training dataset $\mathcal{T}$ which consists of $K\times J$ action sequences for $J$ different actions, each of which with $K$ pre-aligned sequences performed by various individuals. Let $\lambda_i^j$ be the weight value of body joint with label $i$ ($i=1\ldots n$) for the action $j$ ($j=1\ldots J$). Our goal is to find optimal assignment of $\lambda_i^j$ which maximize the similarity error between sequences of different actions and minimize those of same actions. Since the size of the dataset and the alignments of sequences are fixed, this turns out to be an optimization problem on $\lambda_i^j$. Our task is to define a good objective function $f(\lambda_1^j, \lambda_2^j, \dots, \lambda_n)$ for this purpose, and to apply optimization to solve the problem.

\subsection{Weights on Triplets versus Weights on Body Points}
Given a human body model of $n$ points, we could obtain at most $\binom{n}{3}$ triplets, and need to solve a $\binom{n}{3}$ dimensional optimization problem for weight assignment. Even with a simplified human body model of 11 points, this yields a extremely high dimensional ($\binom{11}{3} = 165$ dimensions) problem. On the other hand, the body point triplets are not independent of each other. In fact, adjacent triplets are correlated by their common body points, and the importance of a triplet is also determined by the importance of its three corners (body points). Therefore, instead of using $\binom{n}{3}$ variables for weights of $n$ triplets, we assign $n$ weights $\omega_{1\dots n}$
to the body points $P_{1\dots n}$, where:
\begin{equation}\label{eqn:omega}
\omega_1 + \omega_2 + \ldots + \omega_n = 1.
\end{equation}
The weight of a triplet $\Delta = \left< P_i, P_j, P_k\right>$ are then computed as:
\begin{equation}
\label{eqn:weightAvg}
\lambda_{\Delta} = \frac{\omega_i + \omega_j + \omega_k}{\binom{n-1}{2}}\\
                 = \frac{2(\omega_i + \omega_j + \omega_k)}{(n-1)(n-2)}.
\end{equation}
Note that the definition of $\lambda$ in (\ref{eqn:weightAvg}) ensures that
$\lambda_1 + \lambda_2 + \ldots +\lambda_T = 1$. Using (\ref{eqn:weightAvg}), equation (\ref{eqn:weightedSimilarity}) is rewritten as:
\begin{eqnarray}\label{eqn:weightedSimilarity2}
\mathcal{E}(I_1\rightarrow I_2, J_i\rightarrow J_j) = \frac{2}{(n-1)(n-2)}\nonumber \\
\med_{1 \leq i < j < k \leq n}((\omega_i + \omega_j + \omega_k) \cdot E(\Delta_{i,j,k})),
\end{eqnarray}
By introducing weights $\{\omega_{1\dots n}\}$ to body points, we reduce the high dimensional optimization problem to a lower dimensional, and more tractable problem.

\subsection{Automatic Adjustment of Weights}
Before moving on to the automatic adjustment of weights, we first discuss the similarity score of two pre-aligned sequences.
Given two sequences $A = \{I_{1\dots N}\}$, $B = \{J_{1\dots M}\}$, and the known alignment $\psi : A \rightarrow B$, the similarity of $A$ and $B$ is:
\begin{equation}
\mathscr{S}(A,B)=\sum_{l=1}^{N}{S(l,\psi(l))}
                = N\tau - N \sum_{l=1}^N{\mathcal{E}(I_{l\rightarrow r_1}, J_{\psi(l)\rightarrow r_2})},
\end{equation}
where $r_1$ and $r_2$ are computed reference poses, and $\tau$ is a threshold, which we set as suggested in \cite{Shen2008,Shen2009}. Therefore, the proximate similarity score of $A$ and $B$ is:
\begin{eqnarray}\label{eqn:seqSimilarity}
\bar{\mathscr{S}}(A,B) = N\tau - \frac{2\cdot N}{(n-1)(n-2)} \sum_{l=1}^N \sum_{1 \leq i < j < k \leq n}\nonumber \\
(\omega_i + \omega_j + \omega_k) \cdot E^{l,\psi(l)}(\Delta_{i,j,k}).
\end{eqnarray}
Considering that $N$, $\tau$, $n$ and $E^{l,\psi(l)}(\Delta_{i,j,k})$ are constants given the alignment $\psi$, equation (\ref{eqn:seqSimilarity}) can be further rewritten into a simpler form:
\begin{equation}
\label{eqn:seqSimilarity2}
\bar{\mathscr{S}}(A,B) = a_0 - \sum_{i=1}^{n-1}{a_i \cdot \omega_i},
\end{equation}
where $\{a_i\}$ are constants computed from (\ref{eqn:seqSimilarity}).

Now let us return to our problem of automatic weights assignment for action recognition. As discussed earlier, a good objective function would reflect the intuition that, significant triplets should be assigned higher weights, while trivial triplets should be assigned lower weights. Suppose we have a training dataset $\mathcal{T}$ which consists of $K\times J$ action sequences for $J$ different actions, each of which with $K$ pre-aligned sequences performed by various individuals. $\mathcal{T}_k^j$ is the $k$-th sequence in the group of action $j$, and $\mathcal{R}^j$ is the reference sequence of action $j$. To find the optimal weight assignment for action $j$, we define the objective function as:
\begin{equation}
\label{eqn:weightObjectiveFunc}
f^j(\omega_1, \omega_2, \dots, \omega_{n-1})
= \mathcal{Q}_1 + \alpha\mathcal{Q}_2 - \beta\mathcal{Q}_3 ,
\end{equation}
where $\alpha$ and $\beta$ are non-negative constants and
\begin{equation}
\label{eqn:weightObjectiveFuncTerms1}
\mathcal{Q}_1 = \frac{1}{K}\sum_{k = 1}^{K}{\bar{\mathscr{S}}(\mathcal{R}^j, \mathcal{T}_k^j)},
\end{equation}
\begin{equation}
\mathcal{Q}_2 =
\frac{1}{K}\sum_{k = 1}^{K}{\bar{\mathscr{S}}(\mathcal{R}^j, \mathcal{T}_k^j)^2} - \frac{1}{K^2}\left(\sum_{k = 1}^{K}{\bar{\mathscr{S}}(\mathcal{R}^j, \mathcal{T}_k^j)}\right)^2,
\end{equation}

\begin{equation}
\mathcal{Q}_3 =
\frac{1}{K(J-1)}\sum_{1 \leq i \leq J, i \neq j}\sum_{k = 1}^{K}{\bar{\mathscr{S}}(\mathcal{R}^j, \mathcal{T}_k^i)}.
\end{equation}

The optimal weights for action $j$ are then computed using:
\begin{equation}
\label{eqn:weightSolution}
\left<\omega_1, \omega_2, \dots, \omega_{n-1}\right> = \underset{\omega_1, \omega_2, \dots, \omega_{n-1}}{\operatorname{argmax}}{f^j(\omega_1, \omega_2, \dots, \omega_{n-1})}.
\end{equation}

In this objective function, we use $\mathcal{T}_1^j$ as the reference sequence for action $j$, and the term $\mathcal{Q}_1$ and $\mathcal{Q}_2$ are the mean and variance of similarity scores between $\mathcal{T}_1^j$ and other sequences in the same action. $\mathcal{Q}_3$ is the mean of similarity scores between $\mathcal{T}_1^j$ and all sequences in other different actions. Hence $f^j(\omega_1, \omega_2, \dots, \omega_{n-1})$ achieves high similarity scores for all sequences of same action $j$, and low similarity scores for sequences of different actions. The second term $\mathcal{Q}_2$ may be interpreted as a regularization term to ensure the consistency of sequences in the same group.

As $\mathcal{Q}_1$ and $\mathcal{Q}_3$ are linear functions, and $\mathcal{Q}_2$ is quadratic polynomial, our objective function $f^j(\omega_1, \omega_2, \dots, \omega_{n-1})$ is quadratic polynomial function, and the optimization problem becomes a quadratic programming (QP) problem. There are a variety of methods for solving the QP problem, including interior point, active set, conjugate gradient, etc. In our problem, we adopted the conjugate gradient method, with the initial weight values set to $\left<\frac{1}{n}, \frac{1}{n}, ..., \frac{1}{n} \right>$.

\section{Experiments}
In this section, we apply the proposed weighting based approach to the action recognition problem, and compare its performance with non-weighting methods proposed in \cite{Shen2008,Shen2008-2,Shen2009}. For comparison, we use the same MoCap testing data as in \cite{Shen2008,Shen2008-2,Shen2009}, and build a MoCap training dataset which consists of total of $2 \times 17 \times 5 =  170$ sequences for $5$ actions (walk, jump, golf swing, run, and climb): each action is performed by $2$ subjects, and each instance of action is observed by 17 cameras set up different random locations. The same 11-point human body model in Figure 1 is adopted in the training data. We use the same set of reference sequences for the 5 actions, and align the sequences in the training set against the reference sequences.

To obtain optimal weighting for each action $j$, we first aligned all testing sequences against the reference sequence $\mathcal{R}^j$, and stored the similarity scores of triplets for each pair of matched poses. The objective function $f^j(\omega_1, \omega_2, \dots, \omega_{10})$ is then built based on equation (\ref{eqn:weightObjectiveFunc}), and the computed similarity scores of triplets in the alignments. $f^j(\cdot)$ is a 10-dimensional function, and the weights $\omega_i$ are constrained by
\begin{equation}
\begin{cases}
0 \leq \omega_i \leq 1, i = 1 \ldots 10,\\
\sum_{i=1}^{10}{\omega_i} \leq 1.
\end{cases}
\end{equation}
The optimal weights $\left<\omega_1, \omega_2, \dots, \omega_{10}\right>$ are then searched to maximize $f^j(\cdot)$, with the initialization at $\left<\frac{1}{11}, \frac{1}{11}, \dots, \frac{1}{11}\right>$. The conjugate gradient method is then applied to solve this optimization problem.

\begin{figure}[htb]
\centering
\begin{tabular}{cc}
\includegraphics[width=60mm]{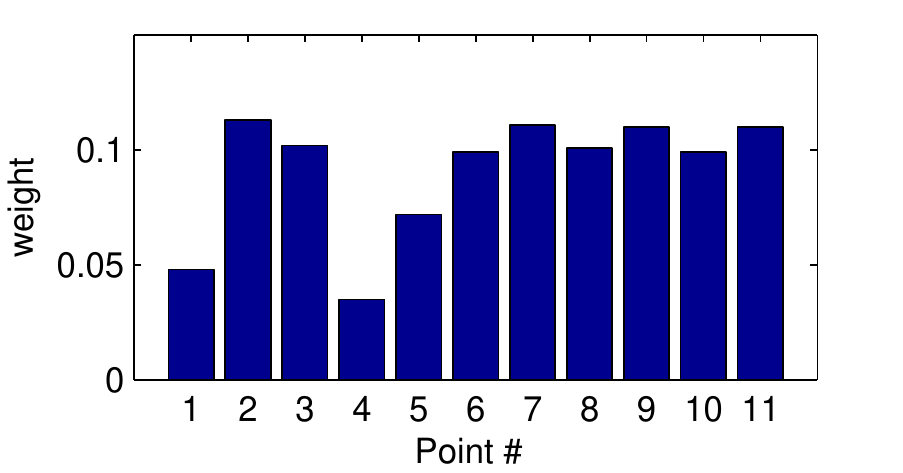} &\includegraphics[width=60mm]{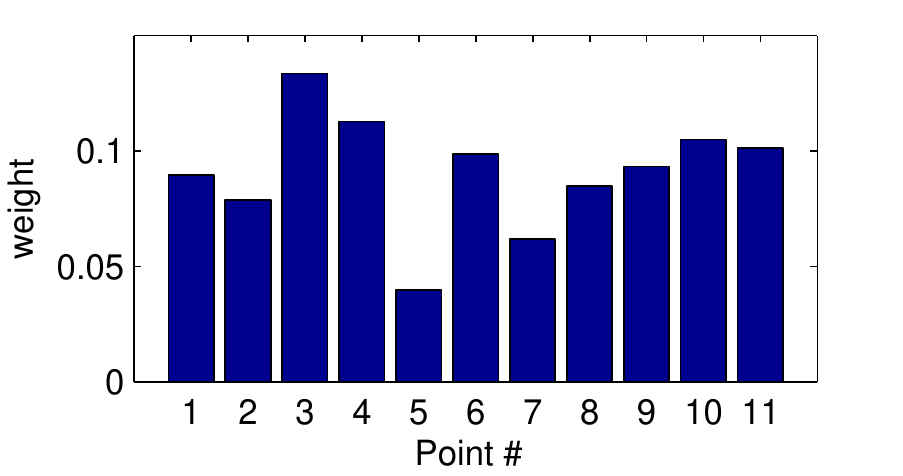} \\
(1) & (2)\\
\includegraphics[width=60mm]{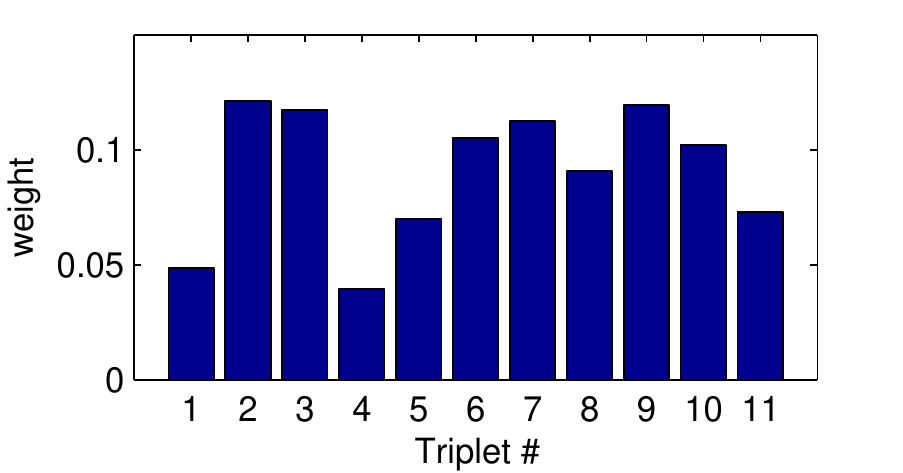} &\includegraphics[width=60mm]{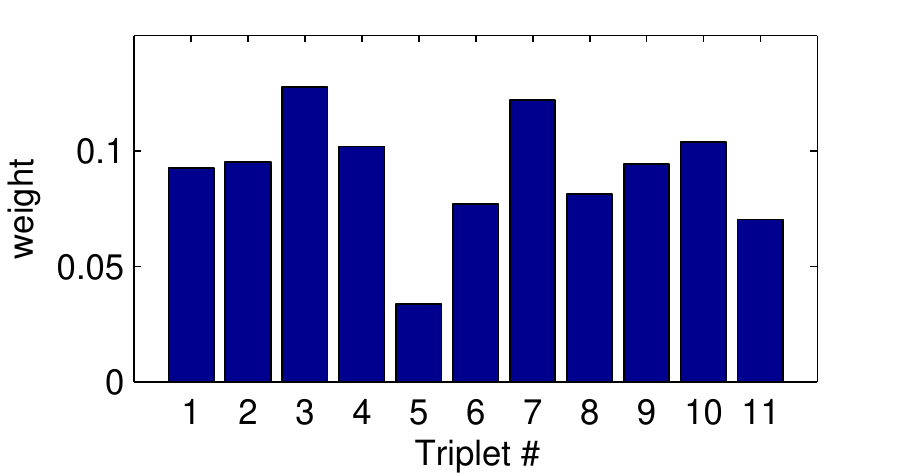}\\
(3) &(4)
\end{tabular}
\caption{Examples of computed weights. (1) and (2) are computed weights for walking and jumping correspondingly based on fundamental ratio invariant; (3) and (4) are computed weights for walking and jumping correspondingly based on eigenvalue equality invariant.}
\label{fig:weightResult}
\end{figure}
After performing the above steps for all the actions, we obtained a set of weights $\mathcal{W}^j$ for each action $j$ in our database. In order to compare our results with existing unweighted methods, we then carried out full comparisons with two methods proposed recently in the literature: action recognition using the fundamental ratios \cite{Shen2008-2}, and the method proposed in \cite{Shen2008,Shen2009}. As these methods behave differently, the objective functions we obtain for estimating weights may contain slightly different sets of coefficients. Figure \ref{fig:weightResult} shows the computed weights for walking and jumping when using for the two methods. Although the weights were slightly different, as shown in the figure similar patterns emerged in terms of significant and trivial triplets: same triplets have relatively high weights in both results.

\begin{table}
\caption{Confusion matrix using eigenvalue equality invariant: Large values on the diagonal entries indicate accuracy.}
\centering
\begin{tabular}{||l||c|c|c|c|c||}
\hline
\multirow{2}{*}{Ground-truth} &\multicolumn{5}{c||}{Recognized as}\\
\cline{2-6}
 &Walk&Jump&Golf Swing&Run&Climb\\
\hline
Walk& 46& 1  & & 1 & 2\\
Jump& 1 & 48 & & &1\\
Golf Swing&1 & & 48 & 1&\\
Run & & 2 & & 48 &\\
Climb& 4 & 1 &1 & & 44\\
\hline
\end{tabular}\label{tab:confusionWeighted}
\end{table}

\begin{table}
\caption{Confusion matrix using fundamental ratios invariant: Large values on the diagonal entries indicate accuracy. }
\centering
\begin{tabular}{||l||c|c|c|c|c||}
\hline
\multirow{2}{*}{Ground-truth} &\multicolumn{5}{c||}{Recognized as}\\
\cline{2-6}
 &Walk&Jump&Golf Swing&Run&Climb\\
\hline
Walk& 45& 1  & 1 & 2 & 1\\
Jump& 2 & 47 & & 1&\\
Golf Swing& &1 & 47 & 1&1\\
Run & 3& 1 & & 45 &1\\
Climb& 4 & 1 & 1&2 & 42\\
\hline
\end{tabular}\label{tab:confusionWeighted2}
\end{table}
We repeated all the action recognition experiments reported in \cite{Shen2008-2} and \cite{Shen2008,Shen2009} using the CMU MoCap data and compared our performance. Results are summarized in Tables \ref{tab:confusionWeighted} and \ref{tab:confusionWeighted2} for the two methods. The overall recognition rate is 93.6\% using a weighted eigenvalue method, and 90.4\% using a weighted fundamental ratios method, which are improved by 2\% and 8.8\% compared to the unweighted cases, respectively.

\section{Conclusion}

We propose a generic method of assigning weights to body points using a small set of training set, in order to improve action recognition from video data. Our method is motivated by the studies in applied perception community on selective processing of visual data for recognition. Our experimental results strongly support our hypothesis that weighting body points differently for different actions leads to significant improvement in performance. Furthermore, since our formulation is based on invariant features, our method shows outstanding performance in the presence of varying camera orientations and parameters. Finally, we believe similar frameworks can be applied to other body models such as silhouette, motion flow, and stick-model.

\bibliographystyle{plain}
\bibliography{foroosh,yuping,refs}

\end{document}